\documentclass[11pt,a4paper]{article}
\usepackage[hyperref]{acl2021}
\usepackage{times}
\usepackage{latexsym}

\usepackage{graphicx}
\usepackage{array}
\usepackage[flushleft]{threeparttable}
\usepackage{multirow}
\usepackage{multicol}
\usepackage{soul}

\usepackage{microtype}

\aclfinalcopy 

\title{ReCAM@IITK at SemEval-2021 Task 4:\\ BERT and ALBERT based Ensemble for Abstract Word Prediction}

\usepackage{soul}

\author{Abhishek Mittal\qquad\ \ \      
  \large{\textbf{Ashutosh Modi}} \\
{Indian Institute of Technology Kanpur (IIT Kanpur)} \\
  {\tt amittal@iitk.ac.in}  \\
  {\tt ashutoshm@cse.iitk.ac.in}  \\
}

\date{}

\begin{document}
\maketitle
\begin{abstract}
This paper describes our system for Task 4 of SemEval-2021: Reading Comprehension of Abstract Meaning (ReCAM). We participated in all subtasks where the main goal was to predict an abstract word missing from a statement. 
We fine-tuned the pre-trained masked language models namely BERT and ALBERT and used an Ensemble of these as our submitted system on Subtask 1 (ReCAM-Imperceptibility) and Subtask 2 (ReCAM-Nonspecificity). For Subtask 3 (ReCAM-Intersection), we submitted the ALBERT model as it gives the best results. We tried multiple approaches and found that Masked Language Modeling(MLM) based approach works the best. 

\end{abstract}

\section{Introduction}
\label{Section 1}
Computers' ability to understand, represent, and express text with abstract meaning is a fundamental problem towards achieving true natural language understanding. In past decades, significant advancement has been achieved in representation learning. SemEval-2021 Task 4 
: Reading Comprehension of Abstract Meaning (ReCAM) \cite{zheng-2021-semeval-task4} explores the ability of machines to understand abstract concepts and proposes to predict abstract words just as humans do while writing article summaries. In the shared task, text passages are provided to read and understand abstract meaning. It consists of three subtasks where the first two subtasks are based on two different definitions of abstractness 1) Imperceptibility \cite{spreen} and 2) Non-specificity \cite{changizi} and the third subtask discusses their intersection. 

Many cloze-style reading comprehension datasets like CNN/Daily Mail \cite{hermann2015teaching} and Children’s Book Test
(CBTest) dataset \cite{hill2016goldilocks} and models  \cite{DBLP:journals/corr/DhingraLCS16, munkhdalai2017reasoning} similar to this task exist,  where a missing word has to be inferred. However, these previous datasets and models have mostly focused on inferring concrete words or concepts like named entities, but this task moves the focus from concreteness to abstractness of words in reading comprehension. This can prove to be quite useful for current ongoing research in the field of abstractive summarization.

We participated in all the three subtasks. We mainly used an Ensemble of BERT and ALBERT as our final model for submission on subtasks 1 and 2. We were ranked 13th on Subtask 1 and 11th on Subtask 2. We submitted the ALBERT model on Subtask 3. All of our code is made publicly available on Github\footnote{\url{https://github.com/amittal151/SemEval-2021-Task4_models}}. We approached this task in two ways. One is a multiple choice Question answering (MCQ) based approach and other a Masked Language Modeling (MLM) approach. Through experiments, we concluded that such tasks are best addressed using a masked language model. 

The rest of the paper is organised as follows. Section \ref{Section 2} describes the problem statement formally and also gives a brief description of the dataset provided by the task  organizers. Section \ref{sec:relatedwork} 
introduces the related work. Section \ref{sec:system} describes our proposed approach and Section \ref{Section 5} gives the experimental details. We enlist our results in Section \ref{Section 6} with a brief error analysis. Finally, we give concluding remarks in Section \ref{Section 7}. 

\begin{figure*}[h]
    \centering
    \includegraphics[width=\textwidth]{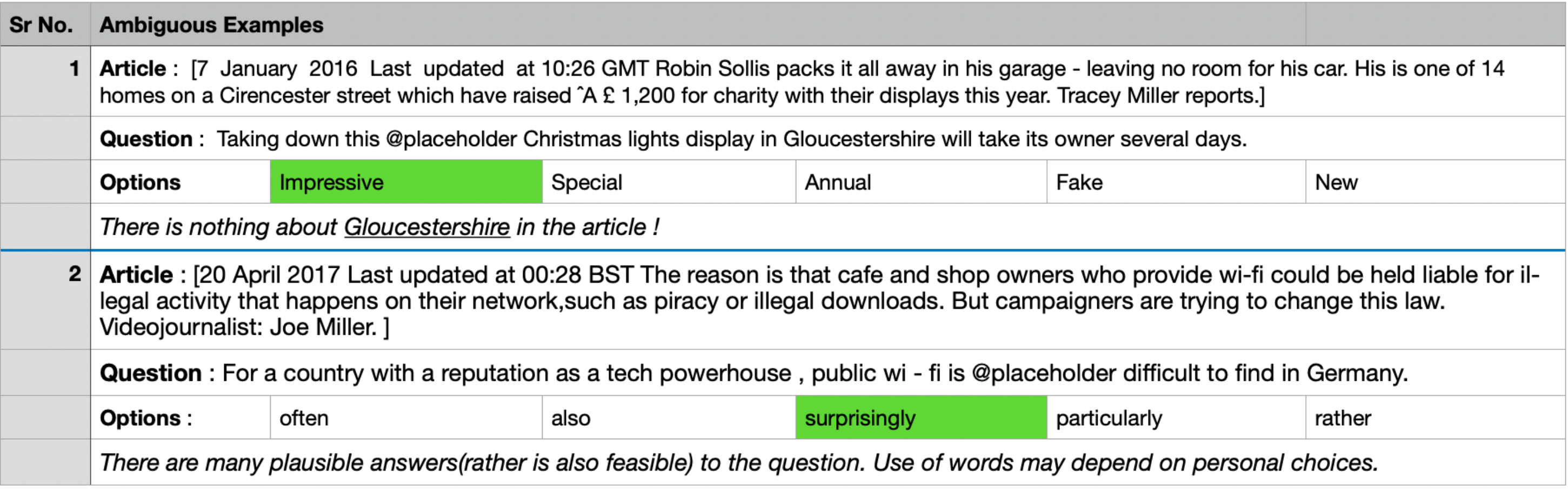}
    \caption{Few ambiguous examples in Subtask 1 Dev set (Option marked in green is the correct answer)}
    \label{fig:ambiguous_ex}
\end{figure*}

\section{Background}
\label{Section 2}

\subsection{Problem Description}
A passage P, a followup question Q with a \textbf{@placeholder} and a list of candidate answer words $W=\{W_{1},W_{2},W_{3},W_{4},W_{5}\}$ are given as an input to the model. The task is to output the correct answer $W_{i}$ from W $(W_{i} \in W)$ by learning the function F such that $W_{i}=F(P,Q,W)$


The first two subtasks focus on the two different definitions of abstractness and the third subtask captures the relationship between the two views of abstractness. The evaluation metric for all three subtasks is the accuracy of the predictions made by the model. The subtasks are enlisted below : 
\begin{enumerate}
    \item \textbf{ReCAM-Imperceptibility} - Abstract words refer to ideas and concepts that are not immediately perceivable by our senses like culture, objective, etc.
    \item \textbf{ReCAM-Nonspecificity} - According to this definition, abstract words refer to holistic terms, e.g., animal, body, etc. 
    \item \textbf{ReCAM-Intersection} - In the third subtask, the system needs to be trained on one definition of abstractness (Imperceptibility) and evaluated on the other (Nonspecificity) and vice-versa.
\end{enumerate}

\begin{table}[h]
    \centering
    \begin{tabular}{|c|c|c|c|}
        \hline
         Task & Training & Dev & Test\\
         \hline
         1 & 3227 & 837 & 2025\\
         \hline
         2 & 3318 & 851 & 2017\\
         \hline
    \end{tabular}
    \caption{Number of examples in dataset}
    \label{tab:data}
\end{table}

\subsection{Data Description}
\label{Section 2.1}

The task organizers have provided training and validation dataset for Subtask 1 and Subtask 2. Each training and validation set example is in the form of a dictionary containing an article, a question and 5 options. One word in the question is missing and is represented by ``@placeholder", and we have to predict the word out of the given 5 options. 

\begin{figure}[h]
    \centering
    \includegraphics[scale=0.25]{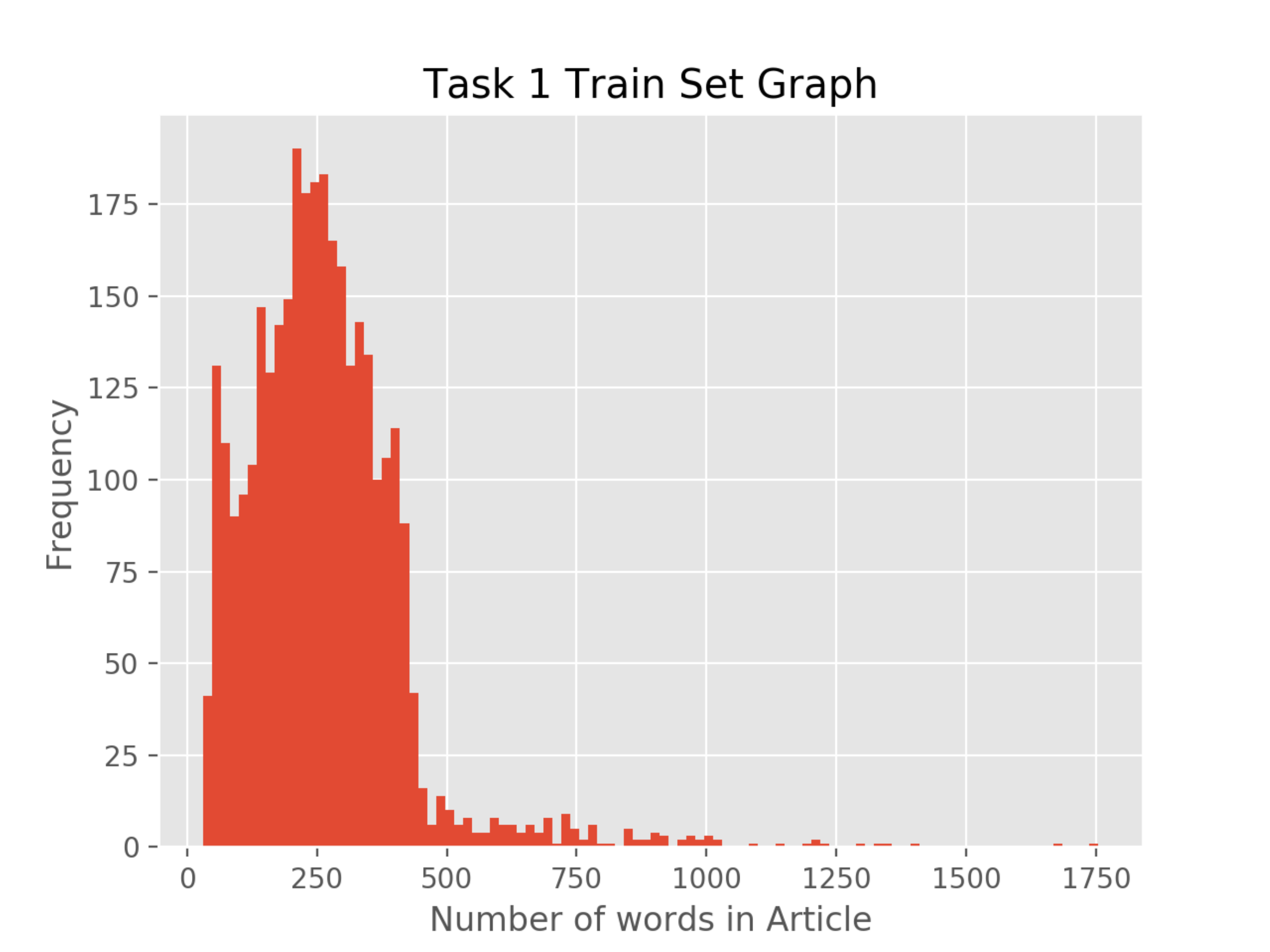}
    \caption{Task 1 Article statistics}
    \label{fig:task1}
\end{figure}

\begin{figure}[h]
    \centering
    \includegraphics[scale=0.25]{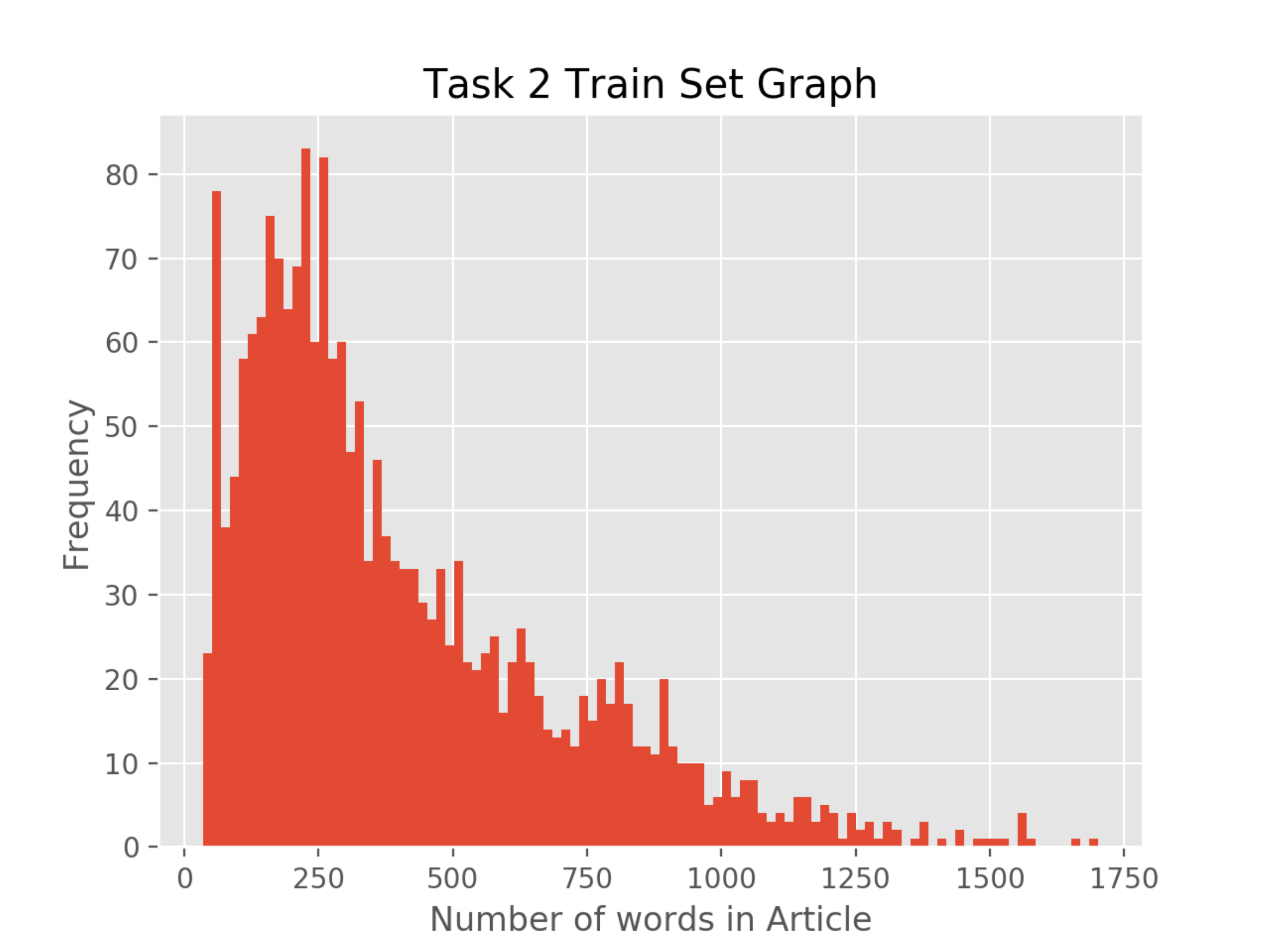}
    \caption{Task 2 Article statistics}
    \label{fig:task2}
\end{figure}

The data set has English news articles and questions are constructed from the summaries of these articles. The data statistics are provided in table \ref{tab:data}. The dataset poses two major challenges. 
Firstly, the passages are quite long. Their distribution is shown in figure \ref{fig:task1} and \ref{fig:task2}. The long article length leads to a loss of context when we truncate the article in a transformer based model due to its max token length limits. Secondly, the dataset contains some ambiguous examples where more than one correct answer could be feasible or the question's context is missing from the article. Some examples are shown in the Figure \ref{fig:ambiguous_ex}. 

\section{Related Work}
\label{sec:relatedwork}

Much work has been done for the prediction of concrete words unlike ours where we need to predict abstract words in reading comprehensions. Gated Attention Reader \cite{DBLP:journals/corr/DhingraLCS16} predicts missing concrete words in CNN/Dailymail datasets with a high accuracy. The attention mechanism plays a crucial role in recognizing which sections of the article are more important to answer the questions. Extracting context from the article is a vital part of the task. This task requires comprehensive natural language understanding, going beyond the meaning of individual words and sentences.  
We explored some of the pre-trained transformer models \cite{vaswani2017attention} as these capture the context better due to the self-attention mechanism.  Moreover, pre-trained models are readily available.

\begin{figure*}[ht]
    \centering
    \includegraphics[scale=0.5]{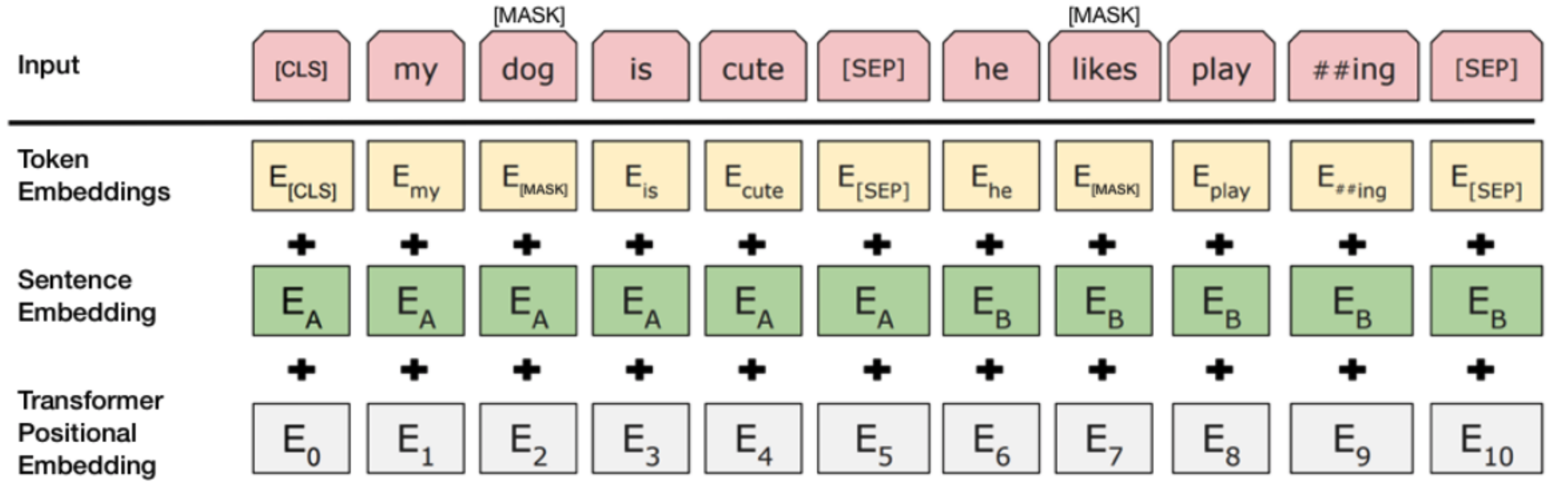}
    \caption{Masked Language Model (Image Src : \cite{Bert}) }
    \label{fig:masked_image}
\end{figure*}

The task is somewhat similar to a multiple choice question answering task. We experimented with the MCQ based approach as mentioned by \citet{Radford2018ImprovingLU} where a linear layer is built over the transformer and the correct answer is predicted by applying softmax over the probabilities of each option. 

One approach to get context from the article is to extract the most relevant sentences to the question with sentence similarity techniques \cite{reimers2019sentencebert}. We experimented with this approach and extracted ``Top-k" sentences that were most semantically similar to the given question from the article.


One of the major challenge in this task is to handle the long length of the article. \citet{pappagari2019hierarchical} discuss the approach of using hierarchical transformers for text classification problem to tackle long passages. BERT is applied to text segments and an LSTM layer or transformer is applied to get document embedding.

Another approach is to model the shared task as a masked language modeling task. 
The transformer based models like BERT 
\cite{Bert} and ALBERT 
\cite{Albert} have been trained via the masked language modeling objective. 
BERT has also been trained on the Next Sentence Prediction task, and ALBERT has been trained on the Sentence Ordering task. In \citet{Albert}, it is mentioned that Sentence Ordering task is a better way to understand the similarity and extracting context from two sentences and thus, ALBERT works better than the BERT model.

\section{System Overview}
\label{sec:system}
We explored multiple models and methodologies. We first experimented using an encoder with an attention based approach. From their results, we observed that an MLM based approach would work better than an MCQ based approach. 


Consequently, we tried BERT and ALBERT models and their Ensemble with the MLM based approach. However, for comparison purposes, we also worked with the MCQ method and its system is described below along with our other approaches.

\subsection{Encoders with Attention}
\label{recurrence}
\subsubsection{Binary Classification with Attention}
\label{Section 4.1.1}

We tried a binary classification based approach where we give each option a score of being a correct answer. Our model consisted of two encoders, followed by a binary classifier. One encoder is for encoding the question and one for encoding the article. First, we feed the question into the question encoder which gives us the context vector of the question. Then, we feed the article along with the hidden weights from the question encoder into the article encoder and apply attention weights over them to find the context of question within the article. Finally, we input an option word, the hidden weights obtained from the article encoder and the context vector from question encoder into the binary classifier layer which gives us the score for the given option. The option word with highest score is predicted as our answer.

\subsubsection{Cosine Similarity of predicted word with options}  
\label{sec:cosine}


In this approach, we use the article and question encoders as described in the first approach to encode the article and question. However, instead of using a binary classifier, we used a decoder layer to predict the missing word.
We used the ``@placeholder'' token's hidden embedding as an input to the decoder layer along with the context vector from question encoder and hidden weights from article encoder. This layer predicts a word from vocabulary which would fit in the place of ``placeholder''. We then compute this word's cosine-similarity with the given 5 options. The most similar option word is predicted as the answer.
This method is similar to an MLM based approach, and the first approach used above is somewhat similar to an MCQ based approach. This method gave slightly better results than the first approach, and thus, it gave us an idea that an MLM model should work better for our subtasks as compared to an MCQ based model.

\subsection{Transformer based Models}
\subsubsection{Multiple choice Question Answering}
\label{sec:mca}
The architecture of this approach is similar to that proposed by \citet{Radford2018ImprovingLU}. We have a linear layer over a transformer model like BERT which takes the embedding of [CLS] token and calculates the cosine similarity of this token with given 5 options. Since, the [CLS] token represents the aggregate sequence representation \cite{Bert}, it encodes the context of article and question together. The input sequence to the model is the concatenation of article and question (where the ``@placeholder" is replaced by an option) delimited with the [SEP] token. On top of this, we have a softmax layer which calculates the score for each given option.

\subsubsection{Masked Language Modeling}
\label{sec:mlm}

We used the transformer models like BERT and ALBERT for masked language modeling since they have been trained via the MLM objective. In this approach, the input sequence to our model is the concatenation of question and article tokens delimited with the [SEP] token where the ``@placeholder" word in the question has been masked. We truncate the article from the end to fit into the maximum token sequence length. Since our task requires context reading from the article, we used different sentence embedding of the transformer models for question and article. An example of input sequence is given in Figure \ref{fig:masked_image}. 
The model's output is a probability vector 
with probability scores of replacing the masked token with any word in the vocabulary. We used the scores computed for the given 5 options and predicted option with the highest score as the correct answer.

We did an ensemble of BERT and ALBERT model predictions by taking the average score for each option predicted by these models. If the scores of Bert model predictions of the 5 options in a given example are B = \{$B_1, B_2, B_3, B_4, B_5$\} and the scores of ALBERT model predictions are A = \{$A_1, A_2, A_3, A_4, A_5$\}, then our Ensemble model gives the scores 
\[ S = \{S_i : S_i = \frac{A_i + B_i}{2} \ \forall i \in\{1,2,3,4,5\} \} \] 
We later also did an Ensemble of two ALBERT models where one is fine-tuned on 
the given subtask, and 
the other one is not. This gave us the best results on Subtask 1 and Subtask 2.  However, we tried this approach in the post-evaluation phase and thus, we did not submit this system on the leaderboard. 

\begin{table}[h]
\centering
\begin{tabular}{ |p{3.2cm}|p{1.2cm}|p{1.2cm}|} 
     \hline
     Model & Task 1 & Task 2\\ 
     \hline
     GAReader (baseline) & 0.251 &  0.243 \\
     \hline
     Binary Classification with Attention & 0.201 & 0.212 \\
     \hline
     Cosine Similarity approach & 0.256 & 0.249 \\
     \hline
    \end{tabular}
    \caption{Results of Encoder Based Approaches on Dev sets (Metric : Accuracy)}
    \label{tab:res}
\end{table}

\begin{table*}[ht]
\centering

\begin{tabular}{|m{0.01\linewidth}|>{\centering\arraybackslash}m{0.35\linewidth} |>{\centering\arraybackslash}m{0.07\linewidth}  |>{\centering\arraybackslash}m{0.07\linewidth} |>{\centering\arraybackslash}m{0.15\linewidth} |>{\centering\arraybackslash}m{0.15\linewidth}|}
\hline
& Model & Task 1 & Task 2 & Task 3 \newline (Train 1, Val 2) & Task 3 \newline(Train 2, Val 1) \\
\hline
1 & Bert (MCQ approach) & 0.195 & 0.202 & 0.201 & 0.198\\
\hline
2 & Bert & 0.6654 & 0.6523 & 0.475 & 0.449  \\
\hline
3 & Bert\_Large - without Article & 0.681 & 0.690 & 0.5347 & 0.5364 \\
\hline
4 & Bert\_Large & 0.7455 & 0.7403 & 0.6451 & 0.5913 \\
\hline
5 & Albert xxlarge - without Article - Not Finetuned  & 0.7011 & 0.7262 & 0.7262 & 0.7011 \\
\hline
6 & Albert xxlarge - Not Finetuned & 0.8172 & 0.8190 & \textbf{0.8190} & \textbf{0.8172} \\
\hline
7 & Albert xxlarge - Finetuned & 0.8291 & 0.8345 & 0.7426 & 0.6977 \\
\hline
8 & Ensemble (4 + 6) & 0.8375 & 0.8401 & 0.7826 & 0.7729 \\
\hline
9 & Ensemble (6 + 7)\textbf{*} & \textbf{0.8685} & \textbf{0.8554} & 0.8143 & 0.8064\\
\hline
\end{tabular}

\caption{Transformer models' results on Dev sets of each task (Metric : Accuracy)}
\begin{flushleft}
    \begin{tablenotes}
      \small 
      \item * Modified System after submission and not submitted in the task
    \end{tablenotes}
\end{flushleft}
\label{tab:transformers}
\end{table*}

\section{Experimental Setup}
\label{Section 5}
Our implementation uses the PyTorch library 
\cite{NEURIPS2019_9015} for deep learning models and the Transformers library by HuggingFace \cite{wolf2020huggingfaces} 
for the pre-trained transformer models and corresponding tokenizers. 

We first experimented with the baseline model, Gated-Attention reader \cite{DBLP:journals/corr/DhingraLCS16} provided by the task organizers. This model did not give good results, as shown in Table \ref{tab:res}.

Then, we experimented with our Encoder based approaches as described in Section \ref{sec:system}. We experimented with various loss functions like NLL loss, MSE and CrossEntropy loss. But, the results were poor for these methods too (Table \ref{tab:res}). 
However, the Cosine similarity based approach (Section \ref{sec:cosine}) performed slightly better than the Binary classification with Attention approach (Section \ref{Section 4.1.1}) indicating that an MLM approach should work better than an MCQ approach.

To verify our claim, we experimented with the BERT Base model with the MCQ approach, which gave quite less accuracy, no better than a random prediction. However, the BERT model with the MLM approach performed way better on all the subtasks. We experimented with both large and small variants of BERT and ALBERT models 
where the large variants performed better as expected. 
We fine-tuned both the models without freezing any layers with Adam optimizer \cite{kingma2017adam}. We fine-tuned the BERT model for 3 epochs and ALBERT model for 1 epoch. We used the learning rate of 5e-5 and a max-sequence length of 256 for both BERT and ALBERT. We also used pre-trained ALBERT model without any fine-tuning in some of our experiments. 

We also experimented with the input sequence to understand and compare the degree of context-reading done in ALBERT and BERT models.  We changed the input sequence to contain only question tokens and then passed this sequence to our models. We then compared these results with the results obtained after passing the complete input sequence containing both article and question tokens. BERT gave an improvement of around 5-6\% after passing the complete input sequence. However, ALBERT 
shows much more improvement of around 11-12\% with the complete sequence. It shows that ALBERT 's training on a Sentence Ordering task is more effective for MLM tasks like ours than the BERT's training on Next Sentence Prediction task. 

We then experimented with the Ensemble of BERT Large and ALBERT xxlarge-v2 model predictions. We experimented by assigning different weights to BERT and ALBERT models and found out that equal weights to both works better. We later did an Ensemble of the fine-tuned ALBERT model with a non fine-tuned ALBERT model. It gave much improved results on Subtask 1 and Subtask 2. 

\begin{table*}[ht]
\small
\centering
\begin{tabular}{|>{\centering\arraybackslash}m{0.1\linewidth}
|m{0.08\linewidth}|
m{0.8\linewidth}|}
\hline
Type & \multicolumn{2}{l}{Example} \\
\hline
\multirow{4}{*}{WC} & \textbf{Article}  & ... "Tennis chose me. It's something I never fell in love with," Tomic told Australia's Channel Seven. "Throughout my career I've given 100\%. I've given also 30\%. But if you balance it out, I think all my career's been around 50\%." .... \\
\cline{2-3}
& \textbf{Question} & Bernard Tomic says he has never " really tried " throughout his tennis career, adding that he has probably been @placeholder at ``around 50\% ". \\
\cline{2-3}
& \textbf{Options}  & (A) held \quad\quad (B) \fcolorbox{red}{white}{aiming} \quad\quad (C) honoured \quad\quad (D) \fcolorbox{blue}{white}{operating} \quad\quad (E) shown \\
\cline{2-3}
& \textbf{Scores} & (A) 16.994 \quad\quad (B) 29.573 \quad\quad (C) 8.331 \quad\quad (D) 18.471 \quad\quad (E) 11.549 \\

\hline

\multirow{4}{*}{WN} & \textbf{Article} &  .... ”These are home games that we have to win," McClaren said. "We are not performing individually and collectively the way that we did up until the Leicester replay. Has that taken too much out of us? I don't know. "We are not getting the rub of the green and we were doing that before. We are not scoring the first goal and we are not scoring goals.... \\
\cline{2-3}
& \textbf{Question} &  Manager Steve McClaren says Derby County 's @placeholder have dropped and he has demanded an immediate response. \\
\cline{2-3}
& \textbf{Options} & (A) \fcolorbox{red}{white}{chances} \quad\quad (B) body \quad\quad (C) \fcolorbox{blue}{white}{standards} \quad\quad (D) side \quad\quad (E) artefacts \\
\cline{2-3}
& \textbf{Scores} & (A) 28.372 \quad\quad (B) 7.169 \quad\quad (C) 27.527 \quad\quad (D) 10.246 \quad\quad (E) 8.395 \\
\hline

\multirow{4}{*}{CC} & \textbf{Article} & Chiriac Inout was found in John Bright Street at about 23:30 GMT on 29 November, one of the coldest nights of the year. Police are investigating after CCTV appeared to show someone searching his pockets while he laid in a loading area behind The Victoria pub. An inquest date is yet to be fixed, the coroner's office confirmed. \\
\cline{2-3}
& \textbf{Question} &  A coroner has named a rough sleeper who may have had property @placeholder before he died in Birmingham city centre . \\
\cline{2-3}
& \textbf{Options} & (A) lost \quad\quad (B) collapsed \quad\quad (C) \fcolorbox{blue}{white}{stolen} \quad\quad (D) delays \quad\quad (E) flowers \\
\cline{2-3}
& \textbf{Scores} & (A) 13.214 \quad\quad (B) 12.342 \quad\quad (C) 27.909 \quad\quad (D) 2.336 \quad\quad (E) 4.510 \\
\hline
\multirow{4}{*}{CN} & \textbf{Article} & .... The Blue Peter team say that Lindsey is safe and on her way back to dry land. Sport Relief said: "Lindsey's Sport Relief challenge was always going to be incredibly hard and zorbing many  miles across the Irish Channel is a huge achievement. ……\\
\cline{2-3}
& \textbf{Question} &  Blue Peter 's Lindsey Russell has ended her attempt to cross the @placeholder between Northern Ireland and Scotland in a giant inflatable barrel for Sport Relief . \\
\cline{2-3}
& \textbf{Options} & (A) gap \quad\quad (B) boundary \quad\quad (C) \fcolorbox{blue}{white}{sea} \quad\quad (D) title \quad\quad (E) difference \\
\cline{2-3}
& \textbf{Scores} & (A) 24.295 \quad\quad (B) 26.728 \quad\quad (C) 26.874 \quad\quad (D) 4.482 \quad\quad (E) 18.486 \\
\hline

\end{tabular}

\caption{Some examples where our model makes mistakes or give correct results on Subtask 2 dev dataset. The options highlighted in blue are the correct answers and options highlighted in red are predicted by our model.}
\begin{flushleft}
    \begin{tablenotes}
      \small 
      \item * WC - Wrong Confident - 27 such examples
      \item * WN - Wrong Confused - 109 such examples
      \item * CC - Correct Confident - 446 such examples
      \item * CN - Correct Confused - 269 such examples
    \end{tablenotes}
\end{flushleft}

\label{tab:examples}
\end{table*}

\section{Results and Analysis}
\label{Section 6} 
The results of all the transformer based approaches are given in Table \ref{tab:transformers}. The recurrence based models did not work and predicted answers with a random probability. We used GloVe \cite{pennington2014glove} vector embeddings  
that are not contextualized, unlike BERT embeddings. Moreover, the task requires some world knowledge since we need to predict an abstract word whose meaning can possibly
be encoded if it is trained on large English corpus. Transformer based models are trained on large corpora and implicitly learn concepts grounded in the world. Hence, transformer based approaches work better in our case.

The BERT model with the MCQ approach uses the embeddings of the `[CLS]' token to predict the correct option. But, it doesn't exploit the position of ``@placeholder'' token and hence it becomes difficult for the model to predict the correct result.

We observe that our Ensemble models work better on Subtask 1 and Subtask 2 
as compared to the BERT and ALBERT models. However, on Subtask 3, the ALBERT model, which is not fine-tuned gives the best results. It owes to the fact that subtasks 1 and 2 differ a lot. If we fine-tune our model on one of the subtask, then it performs worse on the other. 

For our final submission, we submitted our Ensemble model (8) (Table \ref{tab:transformers}) for Subtask 1 and Subtask 2. We also submitted the fine-tuned ALBERT model on these subtasks. In Subtask 1, we are ranked 13th with our Ensemble model (8) with an accuracy of 0.8212 on test set. In Subtask 2, we are ranked 11th with the fine-tuned ALBERT model with an accuracy of 0.8761 on test set. Surprisingly, in Subtask 2, fine-tuned ALBERT model performed better than our Ensemble model on the test set. This is possibly because our ensemble system performed only marginally better than fine-tuned ALBERT model on dev set. 
In Subtask 3, we submitted the non fine-tuned ALBERT model. 

For understanding the mistakes made by our submitted Ensemble system, we analysed the confidence scores of our model's predictions. We performed the analysis of the system on the dev set of Subtask 2. We set a threshold factor (TF) of ``1.4" for deciding between confident and confused predictions. If the confidence score of the model's predicted option is $P$ and the score for the correct option word is $T$, then the model is confident in its predicted answer if the following condition holds : 
\[ P \geq TF * T \] 
It turns out that the model makes 50\% confident predictions out of all the predictions in the dev set. Also, the model makes 20\% wrong predictions confidently, while in 80\% of the wrong predictions, model is confused between two options. In many cases, it is unable to understand the context properly and in a few cases, the model lacks the necessary world knowledge (Table \ref{tab:examples}). In first example in Table \ref{tab:examples}, the model predicts `aiming' as the answer quite confidently. However, `operating' is a more appropriate option due to the context given in the article. This example shows that our system fails to read the context properly in some cases. Also, consider the second example in Table \ref{tab:examples}. Here, the model is confused between two options : `chances' and `standards'. Although, it is mentioned in the article clearly that Derby is performing poor in a past few matches, the model is not confident in predicting `standards' which is the most suitable option here. It implies that our model is generally confused between all the option words that are semantically applicable in the question statement. Consider example 4 from Table \ref{tab:examples}. Here, the model is confused between option words `sea' and `boundary' because both of them fit well into the question. In order to make model more confident on such examples, we need to incorporate world knowledge into our system.  

We also performed a similar post-competition analysis on our Ensemble System (9) (Table \ref{tab:transformers}) and found out that it continues to make similar mistakes. But, in cases like example 1 in Table \ref{tab:examples}, it gives correct results. This is because, we used an ALBERT fine-tuned model instead of BERT fine-tuned model in this system which is better in context-reading as compared to BERT. Thus, this system gave slightly improved results.

\section{Conclusion}
\label{Section 7}
The task of predicting abstract words with context from a question and article is quite novel in itself. We showed that this task can be modelled better as a masked language modeling task rather than multiple choice question answering task. The transformer based approaches worked best, where we used BERT  
and ALBERT models and their ensembles. These models are pretrained models and hence they perform better on our small dataset after fine-tuning. We were able to improve the results of ALBERT model with our Ensemble model on subtasks 1 and 2, but on Subtask 3, the ALBERT model performs better. In future, we shall try to improve our results on Subtask 3. In our current approaches, we haven't used options while training the model. We can try using a pairwise ranking loss function to rank the options according to their scores with a linear layer built on top of transformer models. This will help the model to predict answers more confidently and hence might also improve results. Moreover, we have used the same approach for Subtask 1 and 2. In future, we aim to incorporate common sense knowledge, for example, prototypical knowledge about activities in the form of scripts \cite{modi2014inducing, modi2016event, modi2017modeling, ostermann2018mcscript}, or in the form of semantic networks like ConceptNet \cite{speer2018conceptnet} for tackling two different definitions of abstractness and incorporating some knowledge in the two subtasks.

\bibliographystyle{acl_natbib}
\bibliography{ref}

\newpage

\end{document}